\documentclass{article}

\usepackage{microtype}
\usepackage{graphicx}
\usepackage{subcaption}
\usepackage{booktabs} 

\usepackage{hyperref}

\usepackage[preprint]{icml2026}

\usepackage{amsmath}
\usepackage{amssymb}
\usepackage{mathtools}
\usepackage{amsthm}

\usepackage[capitalize,noabbrev]{cleveref}

\theoremstyle{plain}

\theoremstyle{definition}

\theoremstyle{remark}

\usepackage[textsize=tiny]{todonotes}

\begin{document}

\twocolumn[
  \icmltitle{Environment-Aware Adaptive Pruning with Interleaved Inference Orchestration for Vision-Language-Action Models}



  \icmlsetsymbol{equal}{*}

  \begin{icmlauthorlist}
    \icmlauthor{Yuting Huang}{equal,aff1}
    \icmlauthor{Leilei Ding}{equal,aff1}
    \icmlauthor{Zhipeng Tang}{equal,aff1}
    \icmlauthor{Zenghuan Zhu}{aff1}
    \icmlauthor{Jiajun Deng}{aff1}
    \icmlauthor{Xinrui Lin}{aff1}
    \icmlauthor{Shuo Liu}{aff1}
    \icmlauthor{Haojie Ren}{aff1}
    \icmlauthor{Jianmin Ji}{aff1}
    \icmlauthor{Yanyong Zhang}{aff1}

  \end{icmlauthorlist}
  \icmlaffiliation{aff1}{University of Science and Technology of China, Hefei, China}


  \icmlcorrespondingauthor{Jianmin Ji}{jianmin@ustc.edu.cn}
  \icmlcorrespondingauthor{Yanyong Zhang}{yanyongz@ustc.edu.cn}

  \vskip 0.3in
]



\printAffiliationsAndNotice{\icmlEqualContribution}

\begin{abstract}
While Vision-Language-Action (VLA) models hold promise in embodied intelligence, their large parameter counts lead to substantial inference latency that hinders real-time manipulation, motivating parameter sparsification. However, as the environment evolves during VLA execution, the optimal sparsity patterns change accordingly. Static pruning lacks the adaptability required for environment dynamics, whereas fixed-interval dynamic layer pruning suffers from coarse granularity and high retraining overheads. To bridge this gap, we propose \textbf{EcoVLA}, a training-free, plug-and-play adaptive pruning framework that supports orthogonal combination with existing VLA acceleration methods. EcoVLA comprises two components: \textbf{E}nvironment-aware \textbf{A}daptive \textbf{P}runing (\textbf{EAP}) and \textbf{I}nterleaved \textbf{I}nference \textbf{O}rchestration (\textbf{I\textsuperscript{2}O}). EAP is a lightweight adaptive channel pruning method that incorporates the temporal consistency of the physical environment to update sparsity patterns. I\textsuperscript{2}O leverages the FLOPs bubbles inherent in VLA inference to schedule the pruning method in parallel, ensuring negligible impact on latency. Evaluated on diverse VLA models and benchmarks, EcoVLA delivers state-of-the-art performance, achieving up to 1.60$\times$ speedup with only a 0.4{\%} drop in success rate, and further reaches 2.18$\times$ speedup with only a 0.5{\%} degradation when combined with token pruning. We further validate the effectiveness of EcoVLA on real-world robots.

\end{abstract}

\section{Introduction}


\begin{figure}[h]
    \centering
    \includegraphics[width=0.45\textwidth]{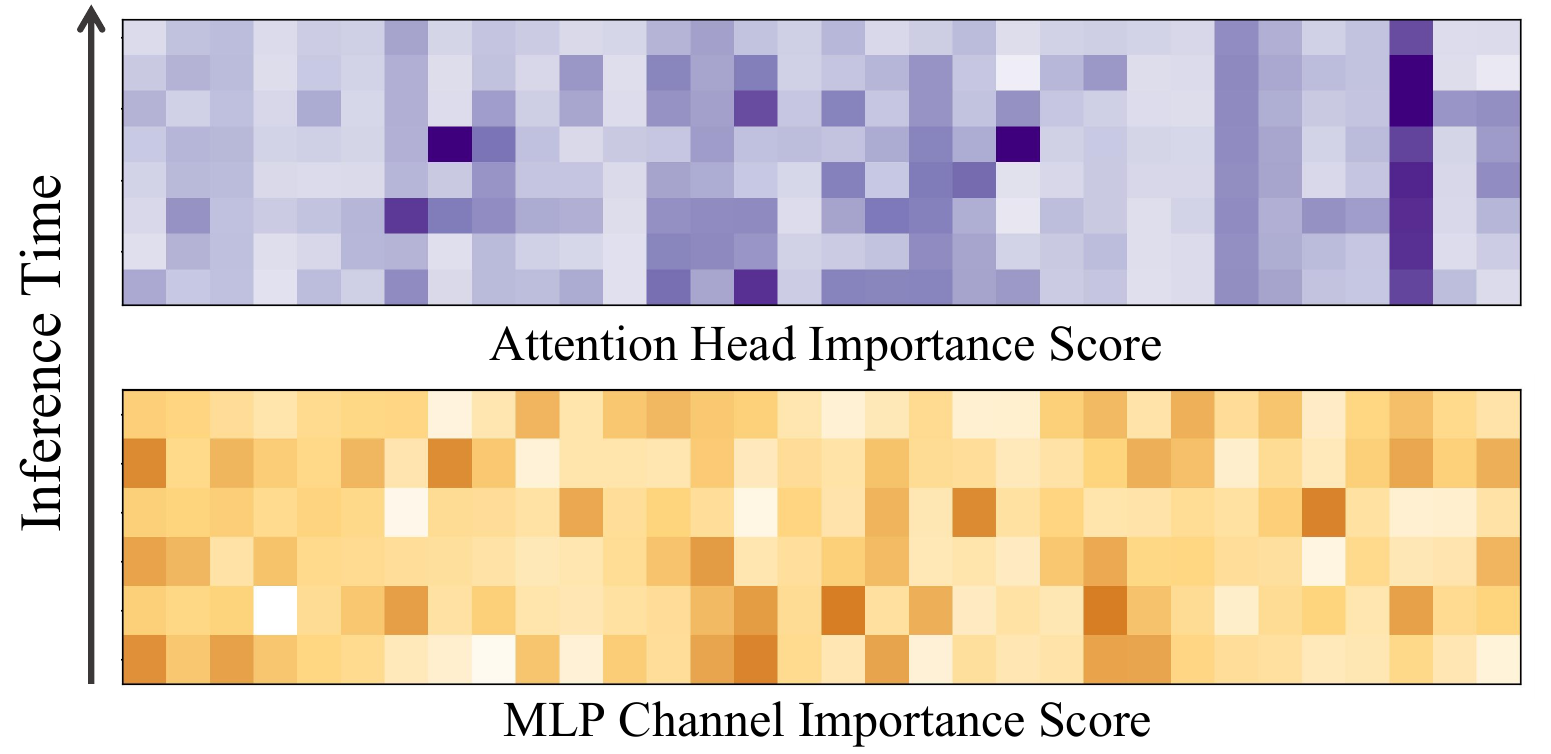}
    \vskip -0.1em
    \caption{During VLA execution, channel importance scores vary dynamically as the environment evolves, causing the optimal sparsity pattern to shift accordingly.}
    \label{fig:channel}
    \vskip -1.2em
\end{figure}

Vision-Language-Action (VLA) models are moving embodied intelligence toward generalization by injecting semantic understanding into robot control~\cite{black2410pi0, li2024cogact, kim2024openvla}. Despite promising real-world results from recent VLA models such as OpenVLA~\cite{kim2024openvla} and $\pi_{0.5}$~\cite{intelligence2504pi0}, inference latency remains the primary bottleneck for real-time control~\cite{yang2025efficientvla}. In this context, token pruning~\cite{wang2025specprune, liu2025vla} has been extensively researched as a method to reduce input size, which decreaces latency by discarding redundant visual tokens at the risk of losing critical semantics with high pruning ratio. While token-level optimization has been thoroughly studied~\cite{shinde2025survey}, comparatively little attention has been paid to VLA model pruning. In this work, we focus on VLA model pruning, an area with less research, to further accelerate inference by reducing redundant model parameters. Typically, a VLA architecture comprises a lightweight action expert and a heavy VLM backbone, which dominates the parameter count and shows high sparsity, making it a prime target for pruning~\cite{chen2025rlrc, jabbour2025don}.




Existing research on VLA model sparsification primarily bifurcates into two directions, static and dynamic pruning. Static pruning, such as RLRC \cite{chen2025rlrc} and GLUESTICK \cite{jabbour2025don}, prunes parameters based on fixed calibration data. However, they fail to adapt to the dynamic evolving task environment (e.g.,transitioning from large-scale navigation to fine-grained local manipulation) where optimal sparsity patterns vary dynamically, as shown in Fig.\ref{fig:channel} \cite{liu2023deja}. Consequently, these static approaches often suffer performance degradation under dynamic sparsity shifts. Moreover, they are heavily shackled by the requirement for massive retraining cycles or a reconstruction process that incurs unsustainable computational costs. To address the inflexibility of static approaches, dynamic pruning like MoLe-VLA~\cite{zhang2025mole} and DeeR-VLA~\cite{yue2024deer} selects layers at fixed intervals based on runtime inputs but suffer from critical drawbacks: the dependency on auxiliary routers incurs extra training and runtime inference overheads, while their coarse layer-level granularity overlooks fine-grained intra-layer redundancy.



To bridge these gaps, a training-free, fine-grained, and environment-aware adaptive model pruning method is urgently needed. However, two major challenges remain. First, VLA models' sparsity patterns evolve with the environment, making real-time computation difficult. Additionally, relying only on instantaneous observations fails to capture the continuous nature of VLA execution. Second, adaptive pruning introduces real-time overhead. While existing methods for LLMs use large-batch inference to amortize overhead across multiple samples~\cite{le2025probe}, VLA models constrained by single-sample streaming bear the pruning overhead individually, directly adding it to the end-to-end inference latency. For frequency-sensitive VLA models, such delays limit the effective policy update rate, inducing robotic stuttering and jittering~\cite{black2025real}.

To address these challenges, we propose \textbf{EcoVLA}, a training-free, plug-and-play pruning framework capable of adapting sparsity patterns via real-time environmental perception while minimizing pruning overhead through non-blocking parallel inference. EcoVLA comprises two core components: \textbf{E}nvironment-aware \textbf{A}daptive \textbf{P}runing (\textbf{EAP}) and \textbf{I}nterleaved \textbf{I}nference \textbf{O}rchestration (\textbf{I\textsuperscript{2}O}).

EAP is a lightweight, environment-aware adaptive structured channel pruning method. First, leveraging visual observations, EAP perceives environmental dynamics to identify variations in sparsity patterns. Crucially, to maintain the temporal consistency essential for stable VLA execution, we incorporate a temporal feature aggregation strategy. By strategically integrating the instantaneous features with historical features, EAP precisely identifies redundant channels for pruning. The continuous update of historical features with the latest features further guarantees the temporal consistency of the sparsity patterns.

I\textsuperscript{2}O replaces the conventional sequential paradigm with a non-blocking parallel paradigm, strategically exploiting FLOPs Bubbles within VLA inference. 
Specifically, I\textsuperscript{2}O orchestrates two parallel streams, comprising an Inference Stream for real-time action generations and a Pruning Stream for pruning execution.
By interleaving pruning computations into the FLOPs bubbles, we maximize overall hardware utilization, effectively masking the pruning overhead to ensure robust real-time streaming control. We evaluate EcoVLA on robotic manipulation across two simulators (LIBERO~\cite{liu2023libero} and SIMPLER~\cite{sun2023simple}) and three VLA models (OpenVLA-OFT~\cite{kim2025fine}, $\pi_{0.5}$~\cite{intelligence2504pi0} and CogACT~\cite{li2024cogact}). EcoVLA achieves $1.6\times$ speedup with only a $0.4\%$ reduction in success rate. Furthermore, by integrating with token pruning methods, EcoVLA boosts the speedup from $1.21\times$ (achieved by FastV with a $50\%$ pruning ratio) to $2.18\times$. Crucially, it recovers the accuracy drop caused by FastV, narrowing the performance gap with the vanilla baseline to just $0.5\%$. To validate our approach beyond simulation, we deploy EcoVLA on a 7-DoF Kinova Gen3 robot, demonstrating its practical acceleration capabilities in real-world scenarios.

\begin{figure*}[h]
    \centering
    \includegraphics[width=0.85\textwidth]{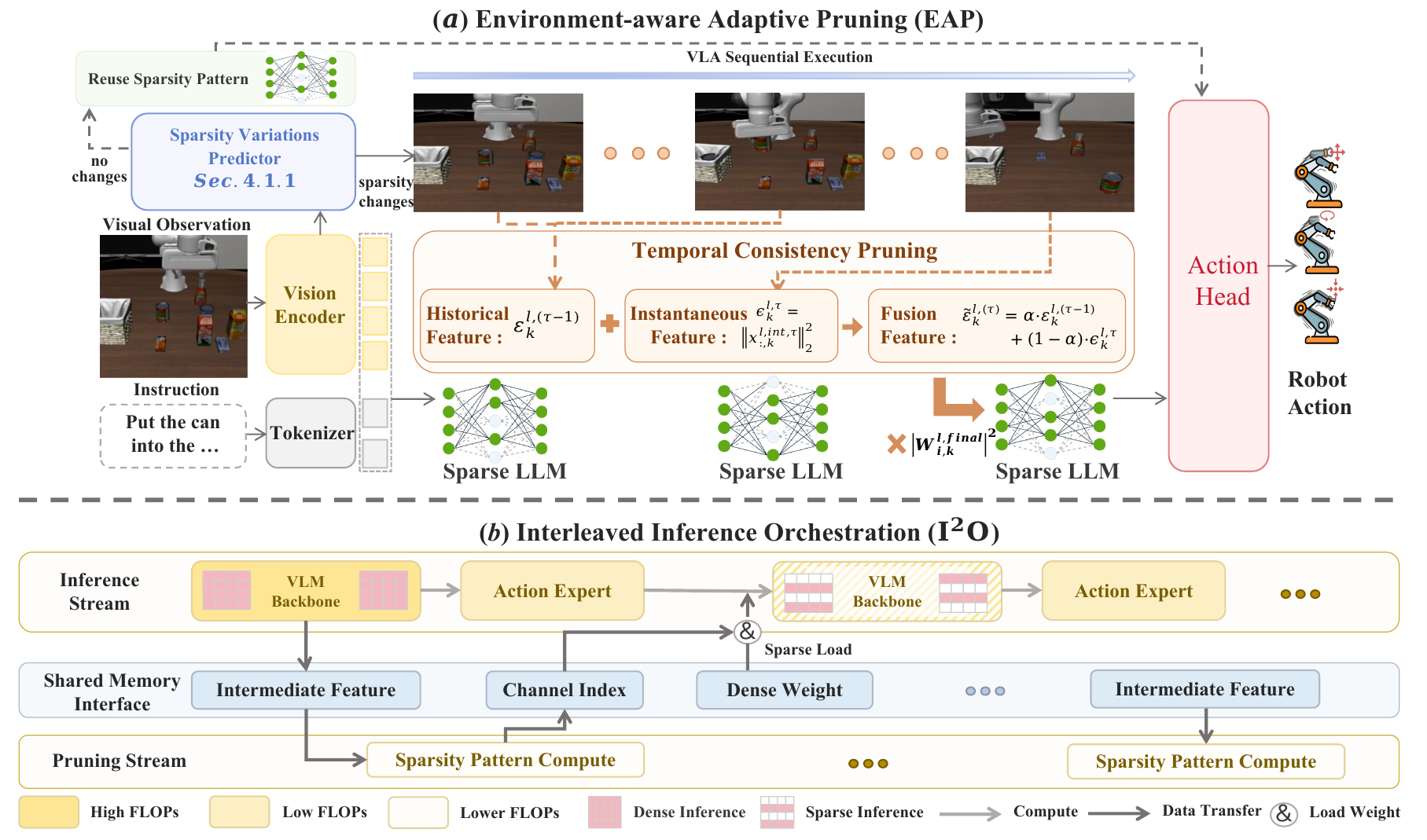}
    \caption{Overall pipeline of EcoVLA. (a) \textbf{E}nvironment-aware \textbf{A}daptive \textbf{P}runing (\textbf{EAP}): EAP is a lightweight, environment-aware method that identifies sparsity variations by perceiving real-time dynamics. Considering the temporal consistency of VLA execution in physical environments, EAP integrates instantaneous features with historical features to jointly compute the sparsity pattern. (b) \textbf{I}nterleaved \textbf{I}nference \textbf{O}rchestration (\textbf{I\textsuperscript{2}O}): I\textsuperscript{2}O interleaves sparsity pattern computation into the inherent FLOPs bubbles within the VLA inference using a non-blocking parallel paradigm.}
    \label{fig:framework1}
    \vskip -1em
\end{figure*}


\section{Related Work}
\textbf{Vision-Language-Action Models.} VLAs extend VLMs by incorporating action modalities for embodied control~\cite{brohan2022rt, black2410pi0}. Despite their efficacy, deployment is constrained by high computational costs~\cite{yang2025efficientvla}. As action heads (e.g., MLPs, diffusion) are lightweight, the VLM backbone remains the dominant computational bottleneck~\cite{zhang2025mole, ma2025running}.

\textbf{Efficient Vision-Language-Action Models. } VLA acceleration typically employs token pruning~\cite{wang2025specprune, liu2025vla} or KV caching~\cite{xu2025vla} to exploit input redundancy. Regarding model sparsification, static methods~\cite{chen2025rlrc, jabbour2025don} prune parameters offline but lack environmental adaptability and require retraining. Conversely, fixed-interval dynamic methods~\cite{yue2024deer, zhang2025mole} offer task-dependence but suffer from high retraining costs and limited transferability. Consequently, a training-free, plug-and-play, and fine-grained adaptive pruning framework is needed.



\section{Preliminaries}
\subsection{Vision-Language-Action Models}
Vision-Language-Action (VLA) models unify visual and linguistic inputs to generate robotic actions, typically consisting of a Vision-Language Model (VLM) backbone and an Action Expert~\cite{kim2025fine, wen2025tinyvla}. While Action Expert architectures vary (e.g., Diffusion-based or Parallel Decoding)~\cite{kim2025fine}, they are lightweight compared to the VLM backbone. Thus, the VLM constitutes the primary bottleneck, rendering its standardized architecture the ideal candidate for pruning.
\subsection{Formulation of VLA Structured Pruning} \label{subsec:Formulation of VLA Structured Pruning}
Let $\pi_{\Theta}$ be a VLA model with Large Language Model (LLM) backbone $\Theta$. We target $\Theta$ for pruning due to its dominant model size. We optimize a structural binary mask $\mathbf{m} \in \{0, 1\}^{|\Theta|}$ (where grouped elements share values) to minimize the divergence between dense and pruned policies:
\begin{equation}
    \mathop{\arg\min}_{\mathbf{m}} \mathcal{L}\left(\pi_{\Theta}, \pi_{\Theta \odot \mathbf{m}}, \mathcal{D}\right) \quad \text{s.t.} \quad \|\mathbf{m}\|_0 = \kappa
    \label{eq:mask_opt}
\end{equation}
where $\odot$ denotes the element-wise product, $\kappa$ is sparsity constraint , $\mathcal{D}$ represents the calibration dataset. The objective $\mathcal{L}$ measures the divergence between models.


To efficiently solve the global optimization problem defined in \eqref{eq:mask_opt}, we decompose the LLM into layer-wise reconstruction paradigm~\cite{an2024fluctuation, le2025probe}. The LLM consists of $L$ blocks. Each block $l$ transforms the input hidden state $\mathbf{X}^l \in \mathbb{R}^{B \times S \times D}$ via a residual mapping:
\begin{equation}
\small      
    \mathbf{X}^{l+1} = \mathbf{X}^l + \mathcal{F}^l(\mathbf{X}^l)
    \label{eq:llm_layer}
\end{equation}
We decompose the block function $\mathcal{F}^l$ into an intermediate transformation and a final linear projection:\begin{equation}
\small      
    \mathcal{F}^l(\mathbf{X}^l) = \mathbf{X}^{l,\text{int}} (\mathbf{W}^{l,\text{final}})^T, \quad  \quad \mathbf{X}^{l,\text{int}} = \mathcal{T}^l(\text{LN}(\mathbf{X}^l))
    \label{eq:hidden_state}
\end{equation}

Here, $\mathcal{T}^l$ is intermediate transformations (e.g.,$\mathbf{W}^K$ or $\mathbf{W}^{\text{up}}$), while $\mathbf{W}^{l,\text{final}}$ represents the final weight matrix (e.g., $\mathbf{W}^O$ or $\mathbf{W}^{\text{down}}$).
For hardware acceleration, structured pruning aligns output channels with input channels, retaining indices $\mathbb{C}^l \subseteq \{1, 2, \dots, C_{\text{in}}\}$:
\begin{equation}
    \begin{aligned}
        & \widetilde{\mathbf{W}}^{l,\text{gate}} = \mathbf{W}^{l,\text{gate}}[\mathbb{C}^l, :], \quad \widetilde{\mathbf{W}}^{l,\text{up}} = \mathbf{W}^{l,\text{up}}[\mathbb{C}^l, :], \\
        & \widetilde{\mathbf{W}}^{l,\text{down}} = \mathbf{W}^{l,\text{down}}[:, \mathbb{C}^l],
        \label{eq:mlp_prune}
    \end{aligned}
\end{equation}
where $\widetilde{\mathbf{W}}^{l,\text{gate}}, \widetilde{\mathbf{W}}^{l,\text{up}} \in \mathbb{R}^{|\mathbb{C}^l| \times C_{\text{out}}}$ and $\widetilde{\mathbf{W}}^{l,\text{down}} \in \mathbb{R}^{C_{\text{out}} \times |\mathbb{C}^l|}$. The notation $|\mathbb{C}^l|$ is the cardinality of $\mathbb{C}^l$. Similarly, in attention blocks, pruning heads removes coupled output channels of $\mathbf{W}^{Q,K,V}$ and input channels of $\mathbf{W}^O$.

\section{Methodology}
In this section, we introduce EcoVLA, the first training-free, plug-and-play adaptive pruning framework for VLA models, as illustrated in Fig.\ref{fig:framework1}. 
We first detail Environment-aware Adaptive Pruning (EAP) in Sec.\ref{sec:eadp}, which begins with a lightweight predictor to identify real-time sparsity variations and subsequently employs a pruning method based on temporal consistency. Next, we introduce Interleaved Inference Orchestration ({I\textsuperscript{2}O}) in Sec.\ref{sec:iio}, a parallel execution paradigm that exploits inference FLOPs bubbles to schedule pruning operations, thereby reducing additional overhead.
\subsection{Environment-aware Adaptive Pruning}
\label{sec:eadp}
\subsubsection{Lightweight Environment-aware Sparsity Variations Predictor}
The optimal sparsity patterns of VLA models evolve dynamically during execution. 
To capture these changes efficiently, we introduce a lightweight environment aware sparsity variations predictor. This module leverages visual feature similarities alongside a temporal context-conditioned trigger, enabling rapid and robust identification of changes.

\textbf{Lightweight Visual Similarity Metric. }
To avoid introducing substantial computational overhead, we discard the popular attention-based semantic similarity~\cite{xu2025vla}, opting instead to leverage the visual features extracted by the VLA visual encoder to compute the similarity between step $t$ and $t-1$. Let $f_t, f_{t-1} \in \mathbb{R}^{N \times D}$ denote the image token features, where $N$ is the number of visual tokens and $D$ is the feature dimension. We define the similarity score as the average token-wise cosine similarity:
\begin{equation}
\small
    s_t = \frac{1}{N} \sum_{i=1}^{N} \frac{f_t^{i} \cdot f_{t-1}^{i}}{\left\| f_t^{i} \right\|_2 \left\| f_{t-1}^{i} \right\|_2 }
    \label{visual_feature_similarity}
\end{equation}
We posit that if the visual features $f_t$ and $f_{t-1}$ exhibit high similarity, the sparsity pattern remains stable between frames. Conversely, significant deviation in visual features implies considerable variation in the sparsity pattern. 


\textbf{Temporal Context–Conditioned Sparsity Trigger. }
In open-world robotic manipulation, the environment is dynamic, causing the distribution of visual feature similarities to evolve over time. Since we rely on these similarities to update sparsity pattern, a static decision criterion becomes brittle under distribution shifts. 
Therefore, we introduce a lightweight Temporal Context-Conditioned Sparsity Trigger, which leverages temporal context to adapt to such shifts. Specifically, we maintain a fixed-size sliding window $\mathcal{H}_t$ storing the similarities of the recent $T$ frames:
\begin{equation}
    \mathcal{H}_t = \{ s_{t-T}, s_{t-T+1}, \dots, s_{t-1} \}
\end{equation}
Based on this temporal context, we adopt a dynamic decision criterion where the sparsity update is triggered if the current similarity drops below the $p$-th quantile of the $\mathcal{H}_t$. The sparsity pattern update policy is formally defined as:
\begin{equation}
    u_t = \mathbb{I}\left( s_t < \text{Quantile}(\mathcal{H}_t, p) \right)
\end{equation}
Here, $p$ serves as a sensitivity hyperparameter, where a higher $p$ enhances responsiveness to subtle changes, whereas lower $p$ prioritizes stability. This mechanism yields self-regulative behavior. During rapid motion, the quantile naturally decreases, suppressing excessive updates to ensure stability. Conversely, in stable phases, the quantile increases, facilitating the sensitive detection of fine-grained variations.

\subsubsection{Temporal Consistency Pruning}
This section details the sparsity pattern computation. Upon triggering a sparsity update at frame $t$, we execute a dense inference to perform the pruning calculation. We first compute the instantaneous features from the current input at frame $t$, which are subsequently aggregated with historical features. The newly computed sparsity pattern is then applied starting from the frame $t+1$ for sparse inference.

To formalize this computation, we denote the update triggered at frame $t$ as the $\tau$-th sparsity update step. Specifically, as the current input reaches block $l$, following the formulation in Sec.\ref{subsec:Formulation of VLA Structured Pruning}, we can compute the current intermediate hidden states $\mathbf{X}^{l,int,\tau}=\mathcal{T}^l(\text{LN}(\mathbf{X}^{l,\tau)}))$. Given that VLA models operate on single-sample streaming inputs, the dimensionality of $\mathbf{X}^{l,\text{int},\tau}$ is defined as $(1, S, C_{in})$. For notational simplicity, we omit the batch dimension in the subsequent formulation. To quantify the activation of the current input across structured channels, we compress the activation along the sequence dimension to obtain the instantaneous feature. Formally, for the $k$-th structured channel, the calculation is performed as follows:
\begin{equation}
    \epsilon_k^{l,\tau} = \sum_{j=1}^{S} \left( \mathbf{X}_{j,k}^{l, \text{int},\tau} \right)^2 = \left\| \mathbf{X}_{:,k}^{l, \text{int},\tau} \right\|_2^2
    \label{eq:channel_activation}
\end{equation}
where $\epsilon_k^{l,\tau}$ represents the instantaneous feature of the $k$-th structured channel given the current input.

However, relying solely on instantaneous features is suboptimal, as the physical execution of VLA models exhibits inherent temporal consistency. This temporal consistency is characterized by smooth, continuous transitions in both physical environments and proprioceptive states across adjacent frames, rather than discrete, abrupt jumps. In light of this, we aggregate the instantaneous feature with the historical feature.
Specifically, we initialize the historical feature $\mathcal{E}^{l,(0)} \in \mathbb{R}^{C_{in}}$ for each block $l$ by applying Eq. \ref{eq:channel_activation} on a calibration dataset. For each subsequent update step $\tau$, we can compute the fused feature by leveraging the previous historical feature $\mathcal{E}^{l,(\tau-1)}$ as a temporal prior:
\begin{equation}
\tilde{\mathcal{E}}_k^{l,\tau} = \alpha \cdot \mathcal{E}_k^{l,(\tau-1)} + (1 - \alpha) \cdot \epsilon_k^{l,\tau}
\end{equation}
where $\alpha \in [0, 1)$ is the temporal inertia parameter. A larger $\alpha$ leads to a more conservative update.
We update the historical feature on full channels using an exponential moving average. Here, the momentum $\lambda$ ensures temporal consistency, providing a stable prior for subsequent steps:
\begin{equation}
    \mathcal{E}^{l,
\tau} = \lambda \cdot \mathcal{E}^{l,(\tau-1)} + (1 - \lambda) \cdot \epsilon^{l,\tau}
\end{equation}

Finally, we can compute the sparsity pattern. Following  PPsp~\cite{le2025probe}, we evaluate the significance of the $k$-th channel in the $l$-th layer based on both weight magnitude and fused feature.
Let $\mathbf{W}^{l, \text{final}} \in \mathbb{R}^{C_{out} \times C_{in}}$ denote the final weight matrix in block $l$. The importance score $\mathcal{S}_k^{l,\tau}$ is formulated as:
\begin{equation}
\mathcal{S}_k^{l,\tau} = \left\| \left\{ \left| W_{i,k}^{l, \text{final}} \right|^2 \cdot \tilde{\mathcal{E}}_k^{l,\tau} \right\}_{i=0}^{C_{out}} \right\|_2
\label{eq:score}
\end{equation}
where $\{ \cdot \}$ denotes the set of elements, and $\mathcal{S}^{l,\tau} \in \mathbb{R}^{C_{in}}$. Crucially, identifying and pruning the $k$-th input channel of $\mathbf{W}^{l, \text{final}}$ (associated with low $\mathcal{S}_k^{l,\tau}$) necessitates the simultaneous removal of the corresponding $k$-th output channel of the  intermediate transformations $\mathcal{T}^l$.  

\subsubsection{Analysis of Computational Cost}
The computational overhead of EAP is primarily divided into visual feature similarity and sparsity pattern computation. For an image represented by $N$ visual tokens with feature dimension $D$ and a block with $C_{in}$ input and $C_{out}$ output channels, the sparsity pattern overhead includes instantaneous feature computation, feature fusion, historical feature updates for $C_{in}$ channels, and importance score computation. Although weight $L_2$-norms are typically pre-computed to reduce online costs to $1 \cdot C_{in}$, the worst-case complexity occurs when weight magnitudes are computed online, leading to the following FLOPs formulation:
\begin{equation}
\small
    \text{FLOPs} \approx 5ND+ 2 S C_{in} + 3 C_{in} C_{out} + 4 C_{in}
\end{equation}
The computational overhead of EAP is marginal, as it primarily consists of element-wise operations that exhibit a negligible footprint compared to the VLA.


\subsection{Interleaved Inference Orchestration}
\label{sec:iio}
In this section, we first analyze the computational characteristics of VLA inference, revealing FLOPs bubbles arising from resource under-utilization. We then introduce \textbf{I}nterleaved \textbf{I}nference \textbf{O}rchestration (\textbf{I\textsuperscript{2}O}), which exploits the complementary resource profiles between the VLM Backbone and Action Expert stages to absorb sparsity pattern computation into these bubbles with minimal overhead (Fig.\ref{fig:framework1}). Finally, we present hardware-efficient implementations to further accelerate both dense and sparse execution.

\subsubsection{FLOPs bubbles of VLA Inference}
In this section, our profiling reveals temporal FLOPs bubbles during VLA inference due to a mismatch between model workload and hardware capacity, providing a opportunity to absorb sparsity pattern computation overhead.

\textbf{The VLM Backbone Stage. }
This stage is dominated by large-scale General Matrix Multiplications (GEMMs) within the transformer layers, rendering it inherently compute-bound. Given the massive model dimensions ($N, K$) and the extensive sequence length ($M$), the arithmetic intensity ($I$) of these operations significantly exceeds the hardware's compute-to-memory ratio:
\begin{equation}
    \frac{\text{FLOPs}}{\text{Bytes}} = \frac{2 \cdot M \cdot N \cdot K}{2 \cdot (NK + MK + MN)} \gg \frac{T_{\text{compute}}}{T_{\text{bandwidth}}}
\end{equation}
where $T_{\text{compute}}$ and $T_{\text{bandwidth}}$ compute throughput and memory bandwidth, respectively. Consequently, the GPU Tensor Cores operate at near-peak saturation, while the memory bandwidth remains largely under-saturated, providing the necessary headroom for concurrent auxiliary tasks.

\textbf{The Action Expert Stage. }
In stark contrast, this stage consists of lightweight MLPs or diffusion-based denoising steps. Operating under the real-time, robotic control streaming (batch size = 1), computational demand falls orders of magnitude below the GPU’s peak parallel processing capacity. In this state, the powerful Tensor Cores remain largely under-utilized, leaving a substantial computational reservoir that can be reclaimed to execute auxiliary tasks.

\subsubsection{Hiding Pruning Overheads via I\textsuperscript{2}O}
 The conventional synchronous pruning stream is serially scheduled either preceding or succeeding the main inference stream, thereby linearly increasing the total latency. To circumvent this bottleneck, EcoVLA introduces \textbf{I}nterleaved \textbf{I}nference \textbf{O}rchestration (\textbf{I\textsuperscript{2}O}) as illustrated in Fig.\ref{fig:framework1}, whose core philosophy is to interleave sparsity pattern computation into the FLOPs bubbles of the VLA pipelines.

Specifically, we decouple the sparsity pattern computation for step $t+1$ from the main dense inference of the current step $t$ by dispatching it to an parallel pruning stream. During the VLM Backbone stage, where the GPU is compute-saturated but memory bandwidth remains undersaturated, the pruning stream concurrently buffers the requisite intermediate activations. Subsequently, as the inference transitions to the Action Expert stage, I\textsuperscript{2}O interleaves the sparsity pattern computation into FLOPs bubbles, effectively utilizing the idle Tensor Cores.  
As a result, I\textsuperscript{2}O fully taps into the GPU's untapped computational potential, achieving a balanced workload distribution across the inference pipeline. By interleaving the sparsity pattern computation into the FLOPs bubbles, our approach avoids intense GPU resource contention, enabling low latency sparsity pattern computation overhead.

Building on this efficient orchestration, we now turn our attention to the latency analysis of I\textsuperscript{2}O. Let $T_{infer}$ denote the VLA inference latency and $T_{prune}$ the overhead. In conventional synchronous approaches, the total latency is additive: $L_{synch} = T_{infer} + T_{prune}$. In I\textsuperscript{2}O, the latency becomes  $L_{I^2O} = T_{infer} + \delta$, 
where $\delta$ accounts for the overhead induced by concurrent execution, such as Streaming Multiprocessor (SM) scheduling costs,  memory bandwidth contention and minor GPU resource competition. Owing to the lightweight design of the adaptive pruning module and its execution within FLOPs bubbles, $\delta$ is minimal. 
Consequently, this orchestration ensures that adaptive pruning is integrated seamlessly without impacting the latency-sensitive VLA control loop, thus maintaining the high-frequency reactivity essential for smooth robotic manipulation.

\subsection{Hardware-efficient Implementation}
In this section, we detail the hardware-level optimizations implemented for both dense and sparse VLA inference to achieve practical inference acceleration.

\subsubsection{Sparse Efficient Kernels}
We apply three kernel-level optimizations targeting memory efficiency and computation throughput in sparse inference.

\textbf{Sparse Linear Transformation Kernel.} In the standard PyTorch implementation, the retained weights are indexed before performing the linear transformation, which incurs additional memory I/O overhead. In contrast, we have implemented a sparse linear transformation Triton kernel that directly loads only the retained weights during computation, avoiding this overhead and improving efficiency.

\textbf{Memory coalescing.}
As described in Sec.\ref{subsec:Formulation of VLA Structured Pruning}, when performing pruning on the $W_{\text{down}}$ weights of the MLP, we actually prune the columns of the weights. In the sparse implementation, this results in indices pointing to non-contiguous memory regions, leading to reduced memory access efficiency. We simply store these matrices in column-major format to enhance memory access locality. Similarly, we also store the output projection weights $W_O$ in the self-attention module in a column-major format.

\textbf{High-Performance Fused Kernels.}
During sparse VLA inference, the MLP computes gate projection, up projection, SiLU activation, and element-wise multiplication. We fuse these four operations into a single Triton kernel that reads the input once and keeps intermediate results in registers, eliminating redundant memory traffic and reducing kernel launches from 4 to 1, yielding nearly 2$\times$ speedup in practice.

\subsubsection{Dense Metric Acceleration} 
In addition to I\textsuperscript{2}O, we apply two memory-level optimizations to accelerate sparsity pattern computation.

\textbf{Allocation-Free Caching.}
As noted in Eq.\ref{eq:score}, computing the sparsity pattern involves weight norms and activation norms, where standard implementations suffer from repeated weight loading and dynamic memory allocation. We address this by pre-computing weight norms as compact vectors (a 99.97\% reduction) and pre-allocating static activation buffers, transforming memory-bound operations into lightweight lookups while eliminating runtime allocations.

\textbf{Batched Metric Computation.} Per-layer sparsity computation incurs many separate kernel launches. We instead stack weight norms and activation buffers across layers into contiguous tensors, enabling kernel fusion that amortizes launch overhead and exposes layer-level parallelism.

\begin{table*}[t]
    \centering
    \scriptsize  
    \renewcommand{\arraystretch}{1.0}
    \makeatletter
    \setlength{\heavyrulewidth}{0.15em} 
    \setlength{\lightrulewidth}{0.12em}
    \setlength{\tabcolsep}{2pt} 
    \vskip 4pt
    \caption{Performance of EcoVLA on OpenVLA-OFT in LIBERO at $25\%$ and $40\%$ pruning ratio.}
    \label{tab:openvla-oft}
    \resizebox{0.85\textwidth}{!}{
        \begin{tabular}{c|ccccc|ccc}
            \toprule
            & \multicolumn{5}{c|}{Success Rate (\%) $\uparrow$} & & & \\
            \cline{2-6}
            \multicolumn{1}{c|}{\raisebox{1.5ex}{Method}} & 
            LIBERO-Spatial & LIBERO-Object & LIBERO-Goal & LIBERO-Long & Average & 
            \raisebox{1.5ex}{FLOPs (T) $\downarrow$} & 
            \raisebox{1.5ex}{Latency (ms) $\downarrow$} & 
            \raisebox{1.5ex}{Speedup $\uparrow$} \\
            
            \midrule
            Vanilla & $97.6$ & $98.4$ & $96.2$ & $94.6$ & $96.7$ & $4.05$ ($100.0\%$) & $143.56$ ($162.78$) & $1.00\times$ \\
            FastV & $96.4$ & $97.2$ & $89.0$ & $94.8$ & $94.4$ & $2.49$ ($61.48\%$) & $118.57$ & $1.21\times$ \\
            VLA-Cache & $96.6$ & $98.6$ & $96.4$ & $94.8$ & $96.6$ & $3.03$ ($74.81\%$) & $148.51$ & $1.10\times$ \\
            \midrule
            \addlinespace[0.1pt]
            \multicolumn{9}{c}{\scriptsize \textit{Pruning Ratio 25\%}} \\
            \addlinespace[-2pt]
            \midrule
            Wanda & $95.8$ & $98.8$ & $87.2$ & $93.2$ & $93.8$ & $3.14$ ($77.53\%$) & $124.32$ & $1.15\times$ \\
            \textbf{Ours} & $97.4$ & $98.8$ & $94.6$ & $96.4$ & $96.8$ & $3.23$ ($79.75\%$) & $113.98$ & $1.26\times$ \\
            \textbf{FastV + Ours} & $96.6$ & $98.2$ & $94.0$ & $96.0$ & $96.2$ & $1.96$ ($48.39\%$) & $65.85$ & $2.18\times$ \\
            \textbf{VLA-Cache + Ours} & $96.8$ & $98.4$ & $92.4$ & $94.2$ & $95.5$ & $2.43$ ($60.00\%$) & $121.24$ & $1.34\times$ \\
               
            \midrule
            
            \addlinespace[-2pt] 
            \multicolumn{9}{c}{\scriptsize \textit{Pruning Ratio 40\%}} \\
            \addlinespace[-2.5pt] 
            \midrule
            Wanda & $89.2$ & $98.4$ & $77.0$ & $90.6$ & $88.8$ & $2.57$ ($63.46\%$) & $106.47$ & $1.35\times$ \\
            \textbf{Ours} &$ 95.6$ & $97.8$ & $89.4$ & $93.2$ &$94.0$ & $2.74$ ($67.65\%$) & $101.58$ & $1.41\times$ \\
            \textbf{FastV + Ours} & $94.4$ & $97.8$ & $85.8 $&$ 93.4 $& $92.9$ & $1.64$ ($40.49\%$) & $61.16$ & $2.35\times$ \\
            \textbf{VLA-Cache + Ours} & $93.0$ & $98.2$ & $89.6$ & $93.4$ & $93.6$ & $2.03 $($50.12\%$) & $108.48$ & $1.50\times$ \\
            
            \bottomrule
        \end{tabular}
    }

    \vskip 4pt

    \caption{Performance of EcoVLA on $\pi_{0.5}$ in LIBERO at $25\%$ and $37.5\%$ pruning ratio.}
    \label{tab:pi05}
    
    \begin{small}

    \resizebox{0.85\textwidth}{!}{%
    
        \begin{tabular}{cc|ccccc|ccc}
        \toprule 
        
        & & \multicolumn{5}{c|}{Success Rate (\%) $\uparrow$} & & & \\
        
        \cline{3-7}
        
        \multicolumn{1}{c}{\raisebox{1.5ex}{Method}} & 
        \raisebox{1.5ex}{Sparsity} & 
        LIBERO-Spatial & LIBERO-Object & LIBERO-Goal & LIBERO-Long & Average & 
        \raisebox{1.5ex}{FLOPs (T) $\downarrow$} & 
        \raisebox{1.5ex}{Latency (ms) $\downarrow$} & 
        \raisebox{1.5ex}{Speedup $\uparrow$} \\
        
        \midrule 
        
        Vanilla & $0$\% & $98.8$ & $98.2 $& $98.0$ & $92.4$ & $96.9$ & $1.99$ ($100.0\%$) & $81.94$ & $1.00\times$ \\
        
        \midrule 
        \raisebox{-1.5ex}[0pt][0pt]{\textbf{Ours}} & $25.0\%$ & $98.2$ & $98.6$ & $98.4$ & $91.6$ & $96.7$ & $1.64$ ($82.41\%$) & $62.66$ & $1.31\times$ \\
        
                      & $37.5$\% & $97.8$ & $98.4$ &$ 96.8$ & $87.0$ & $95.0$ & $1.47$ ($73.87\%$) & $55.98$ & $1.46\times$ \\
        
        \bottomrule
        \end{tabular}
    }
    \end{small}

    \vskip 4pt
    \caption{Performance of EcoVLA on CogACT in SIMPLER at $25\%$ and $40\%$ pruning ratio.}
    \label{tab:cogact}

    \resizebox{0.85\textwidth}{!}{%
        \begin{tabular}{c|cc|ccccc|ccc} 
        \toprule
        
        & & & \multicolumn{5}{c|}{Success Rate (\%) $\uparrow$} & & & \\
        \cline{4-8}
        
        \raisebox{1.5ex}{SIMPLER} & 
        \raisebox{1.5ex}{Method} & 
        \raisebox{1.5ex}{Sparsity} & 
        Pick Coke & Move Near & Open/Close & Open Top & Average & 
        \raisebox{1.5ex}{FLOPs(T)$\downarrow$} & 
        \raisebox{1.5ex}{Latency (ms) $\downarrow$} & 
        \raisebox{1.5ex}{Speedup $\uparrow$} \\
        
        \midrule

                                 & Vanilla & $0$    & $93.3$ & $83.8$ & $74.5$ & $41.7$ & $73.3$ & $1.81$ $(100\%)$ & $104.16$ & $1.00\times$ \\
        Visual Matching & \textbf{ours}     & $25\%$ & $95.0$ & $82.1$ & $70.8$ & $38.9$ & $71.7$ & $1.45$ $(80.11\%)$ & $72.65$ & $1.44\times$ \\
                                 & \textbf{ours}     & $40\%$ & $93.0$ & $85.4$ & $73.5$ & $42.6$ & $73.6$ & $1.25$ ($69.06\%)$ & $66.43$ & $1.57\times$ \\
        
        \midrule

                                 & Vanilla & $0$    & $88.7$ & $76.8$ & $26.7$ & $51.9$ & $61.0$ & $1.81$ $(100.0\%)$ & $105.87$ & $1.00\times$ \\
        Variant Aggregation & \textbf{ours}     & $25$\% & $85.9$ & $75.3$ & $27.2$ & $46.0$ & $58.6$ & $1.47$ $(81.22\%)$ & $73.98$ & $1.43\times$ \\
                                 & \textbf{ours}     & $40\%$ & $86.1$ & $74.3$ & $33.1$ & $48.7$ & $60.6$ & $1.28$ $(70.72\%)$ & $66.25$ & $1.60\times$ \\
        
        \bottomrule
        \end{tabular}
    }
    \vskip -10pt
\end{table*}

\begin{table}[ht]
    \centering
    \footnotesize
    \makeatletter
    \setlength{\heavyrulewidth}{0.12em} 
    \setlength{\lightrulewidth}{0.12em}
    \caption{Performance of EcoVLA on $\pi_{0.5}$ on a real-world robot.}
    \label{tab:realrobot}
    \begin{tabular}{c|c|c|c|c} 
    \toprule
    Method      & Task1 & Task2 & Task3 & Latency(ms) \\ \midrule
    baseline    & $12/20$ & $18/20$ & $16/20$ & $86.08$ \\
    Ours        & $12/20$ & $16/20$ & $15/20$ & $68.40$ \\ \bottomrule
    \end{tabular}
\end{table}

\section{Experiments}
\subsection{Experimental Settings}

\textbf{Baselines. }
To validate the generalizability of EcoVLA, we evaluate it across diverse VLA architectures. In simulation, we conduct experiments on open-source VLA models, including OpenVLA-OFT~\cite{kim2025fine}, $\pi_{0.5}$~\cite{intelligence2504pi0}, and CogACT~\cite{li2024cogact}. For fair comparison, we benchmark EcoVLA against Wanda~\cite{sun2023simple}, a mainstream static pruning method. Beyond standalone improvements, we demonstrate that EcoVLA is broadly compatible and can be stacked with other acceleration techniques. Experiments combining EcoVLA with FastV~\cite{chen2024image} and VLA-Cache~\cite{xu2025vla} yield substantial additional speedups with negligible performance degradation. For real-world evaluation, we fine-tune $\pi_{0.5}$ for deployment on a 7-DoF Kinova Gen3 robotic arm. The model is trained for $10,000$ steps with default parameters. See App.\ref{appendix:Baseline} for further details.


\textbf{Evaluation Protocol. }To ensure a comprehensive comparison, we adopt different pruning ratios. Specifically, for OpenVLA-OFT~\cite{kim2025fine} and CogACT~\cite{li2024cogact}, we perform evaluations at pruning ratios of $25\%$ and $40\%$, while for $\pi_{0.5}$~\cite{intelligence2504pi0}, we evaluate at $25\%$ and $37.5\%$ to accommodate the architecture. The evaluation metrics primarily include task success rate$(\%)$, inference latency$(ms)$, and FLOPs$(T)$.

\textbf{Implementation Details. }We follow the original experimental settings of FastV~\cite{chen2024image} and VLA-Cache~\cite{xu2025vla}. All experiments conducted on an NVIDIA RTX 3090 GPU, and latency is measured following VLA-Cache. Since VLA-Cache is incompatible with \texttt{FlashAttention}~\cite{dao2022flashattention}, we use eager \texttt{LlamaAttention} when integrating EcoVLA with VLA-Cache; otherwise, FlashAttention is enabled by default. More hyperparameter details are provided in the App.\ref{appendix:Implementation}.

\begin{figure}[h]

    \centering
    \includegraphics[width=0.4\textwidth]{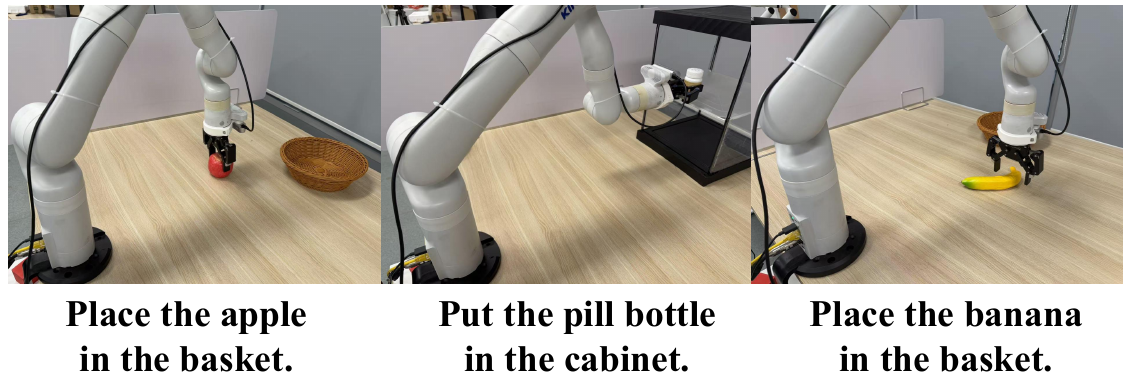}
    \caption{Robot Manipulation on Kinova Gen3 Platform.}
    \label{fig:real-robot}
    \vskip -1em
\end{figure}

\textbf{Benchmarks. }We evaluate EcoVLA on two simulators and three real-robot tasks. The LIBERO~\cite{liu2023libero} benchmark is designed to evaluate robotic manipulation capabilities across four task suites: LIBERO-Spatial, LIBERO-Object, LIBERO-Goal, and LIBERO-Long. Each task suite examines different capabilities of VLA. The SIMPLER~\cite{li2024evaluating} simulation environment provides two evaluation settings, including Visual Matching and Variant Aggregation. We evaluate four tasks on the Google Robot arm: 1) \textit{Pick Coke Can (PickCan)}, 2) \textit{Move Near (MoveNear)}, 3) \textit{Open/Close Drawer (Drawer)}, and 4) \textit{Open Top Drawer and Place Apple (DrawerApple)}. To assess EcoVLA in real-world settings, we conduct experiments across three tasks (shown in Fig.\ref{fig:real-robot}): \textit{Place the apple in the basket}, \textit{Put the pill bottle in the cabinet}, and \textit{Place the banana in the basket}. More details are provided in the App.\ref{app:bench}.



\begin{figure}[h]

    \centering
    \includegraphics[width=0.4\textwidth]{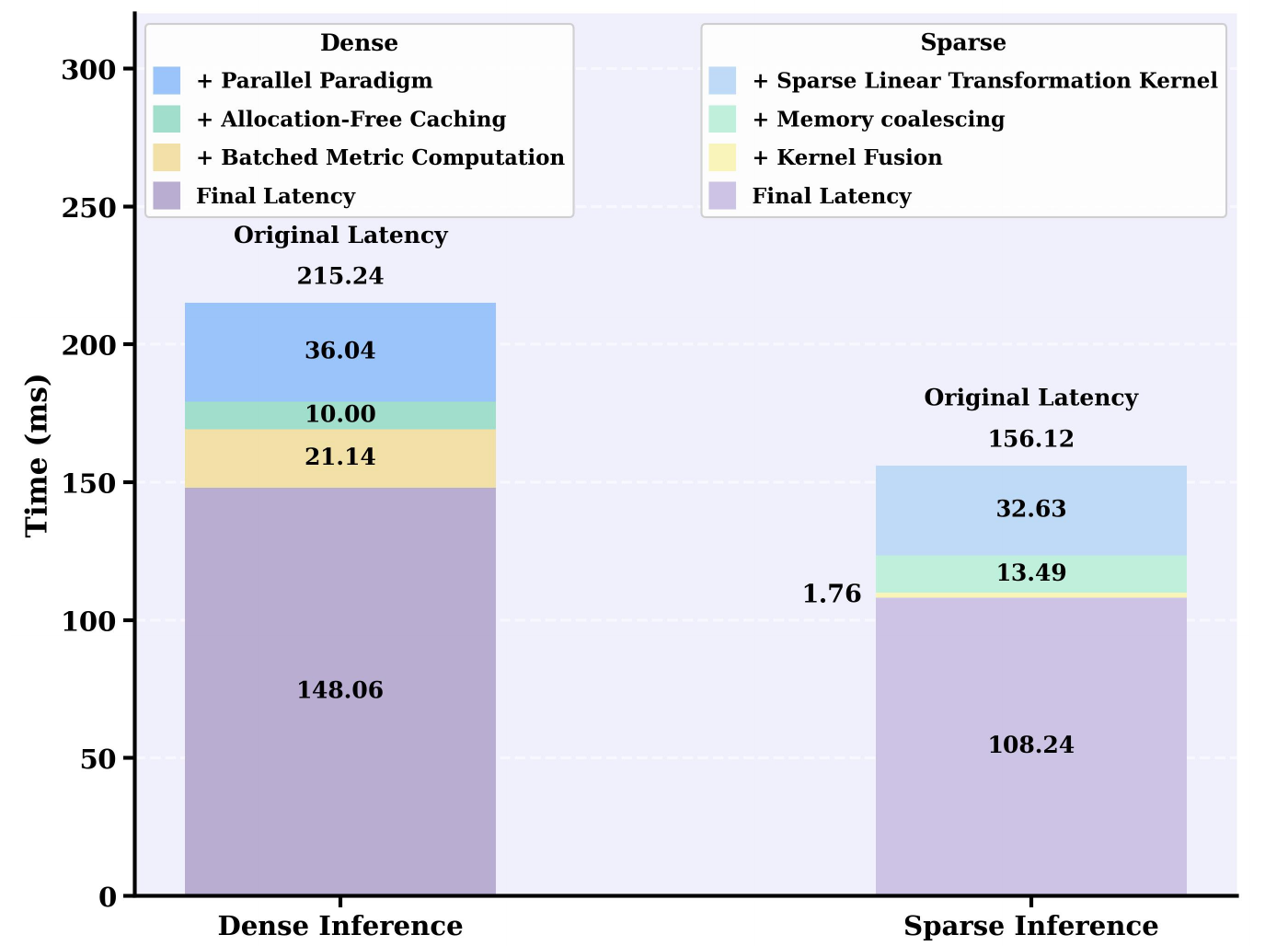}
    \caption{Acceleration breakdown for dense and sparse inference.}
    \label{fig:roofline}
    \vskip -1.5em
\end{figure}

\subsection{Main Results}
\textbf{Results on OpenVLA-OFT.}
Using the LIBERO benchmark, we evaluate EcoVLA on OpenVLA-OFT and show its compatibility by combining it orthogonally with acceleration techniques like FastV and VLA-Cache.
Tab.\ref{tab:openvla-oft} shows that EcoVLA achieves $1.26\times$ and $1.41\times$ speedups at $25\%$ and $40\%$ pruning ratios, with success rate losses of only $0.35\%$ and $2.8\%$, respectively.
The robustness of EcoVLA is particularly pronounced on the pruning-sensitive LIBERO-Goal benchmark, significantly outperforming the mainstream method Wanda. Specifically, EcoVLA establishes a commanding lead of $7.4\%$ and $12.4\%$ over Wanda at the respective pruning ratios.
Beyond this robustness, EcoVLA achieves lower latency than Wanda despite a additional  pruning overhead, as I\textsuperscript{2}O conceals the pruning cost within FLOPs bubbles. Combined with our hardware-efficient design, this enables EcoVLA to deliver distinctly higher wall-clock speedups over Wanda.

\begin{figure}[h]

    \centering
    \includegraphics[width=0.45\textwidth]{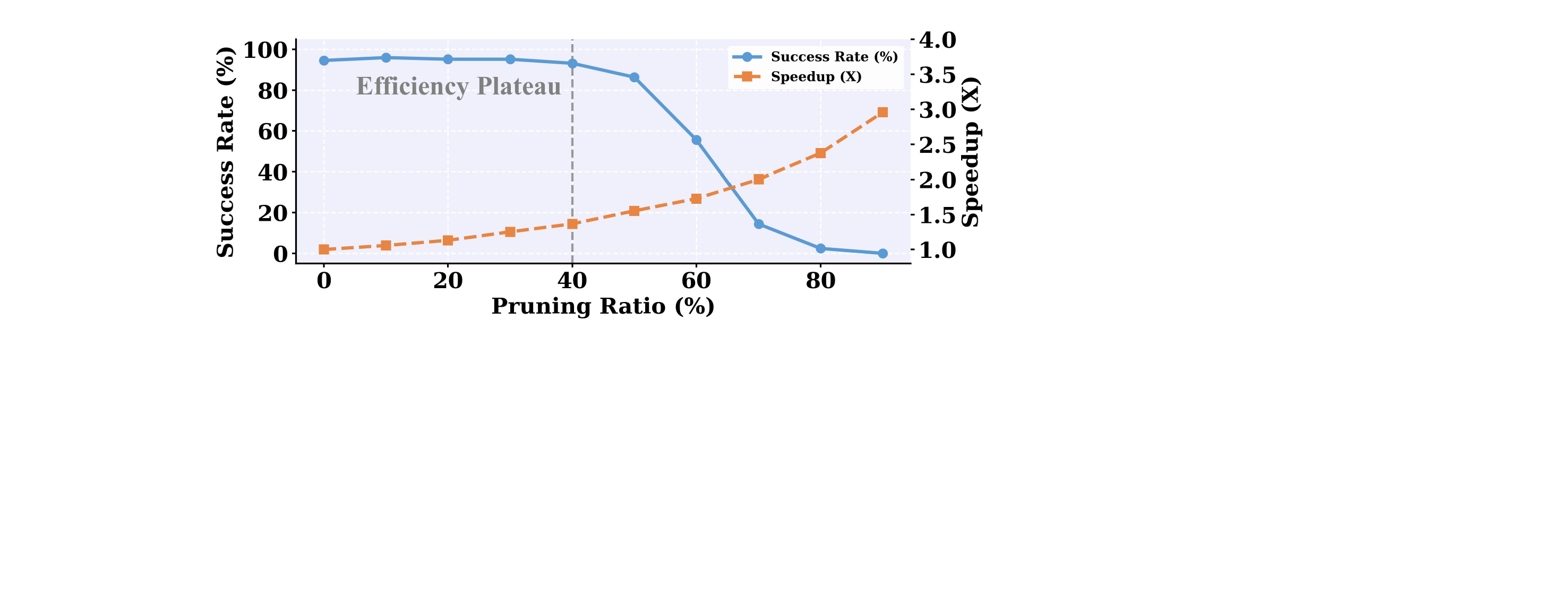}
    \caption{Trade-off between Success Rate and Latency.}
    \label{fig:efficient}
    \vskip -1.2em
\end{figure}

\begin{table}[ht]
    \centering
    \footnotesize
    \makeatletter
    \setlength{\heavyrulewidth}{0.12em} 
    \setlength{\lightrulewidth}{0.12em}
    \caption{Overhead of Pruning Stream.}
    \label{tab:latency_analysis}
    \begin{tabular}{c|c} 
    \toprule
    \textbf{Execution Method} & \textbf{Latency (ms)} \\ \midrule
    Normal VLA Inference    & 143.56 \\
    I\textsuperscript{2}O    & 148.06 \\ \bottomrule
    \end{tabular}
    \vskip -1.2em
\end{table}

Our results empirically validate the broad compatibility of EcoVLA. 
When combined with FastV ($50\%$ token pruning), it achieves a $2.18\times$ speedup with a negligible $0.5\%$ loss at $25\%$ sparsity, and reaches a $2.35\times$ speedup at $40\%$.
When combined with VLA-cache, it achieves a $1.34\times$ speedup with a negligible $0.5\%$ loss at $25\%$ sparsity, and reaches a $2.35\times$ speedup at $40\%$.
Notably, these combinations deliver significantly higher speedups compared to the standalone baselines. While generally incurring only a slight decline in success rate, intriguingly, we observe a performance improvement in the FastV combination. We attribute this phenomenon to the regularization effect induced by our precise, adapti  model sparsification \cite{jin2022pruning}.

\textbf{Results on $\pi_{0.5}$. }To demonstrate the cross-model generalizability of EcoVLA, we conduct experiments on the state-of-the-art VLA model, $\pi_{0.5}$, as shown in Tab.\ref{tab:pi05}. 
At $25\%$ and $37.5\%$ pruning ratios, EcoVLA achieves $1.31\times$ and $1.46\times$ speedups with marginal accuracy drops of $0.2\%$ and $1.85\%$, respectively. Notably, for the LIBERO-Object, we observe a $0.2\%$ accuracy improvement at the $37.5\%$ pruning ratio. This observation corroborates the regularization benefit discussed earlier, suggesting that selective pruning effectively filters out noise from redundant parameters. Most significantly, despite $\pi_{0.5}$’s inherently efficient structure and low latency ($81.94$ ms), EcoVLA still manages to extract a substantial $1.46\times$ acceleration. This capability to further accelerate an already fast model redefines the standards for efficient real-time deployment.


\textbf{Results on CogACT. }We evaluate EcoVLA's generalization on CogACT within SIMPLER, as summarized in Tab.\ref{tab:cogact}. In the Visual Matching setting, EcoVLA achieves a $1.57\times$ speedup at a $40\%$ pruning ratio with a success rate drop of only $0.3\%$. Similarly, in the Variant Aggregation setting, it reaches a $1.6\times$ speedup at $40\%$ pruning while incurring a marginal accuracy reduction of $0.6\%$.

\begin{figure}[h]
    \centering
    \includegraphics[width=0.4\textwidth]{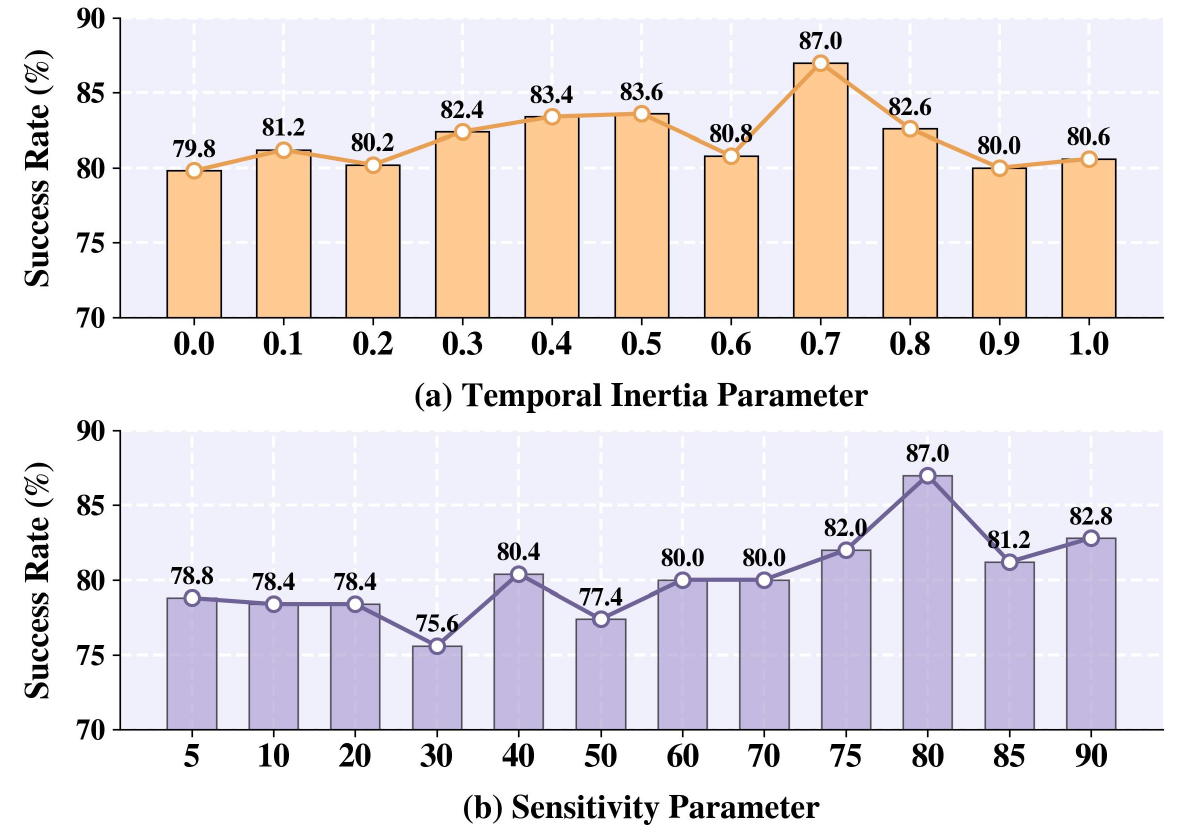}
    \caption{Impact on Hyperparameters $\alpha$ and $p$.}
    \label{fig:para}
    \vskip -1.5em
\end{figure}

\textbf{Results on Real Robot. }We evaluate real-world performance on a physical 7-DoF Kinova Gen3 arm controlled by $\pi_{0.5}$. As shown in Tab.\ref{fig:real-robot}, our approach delivers a $1.26\times$ wall-clock speedup at the cost of minor performance loss, underscoring its viability for real-robot deployment. We further analyze the underlying reasons in App.\ref{app:realrobot}.



\subsection{More Results}

\textbf{Ablation Study on Acceleration. }
Fig.\ref{fig:roofline} details the acceleration breakdown. For dense inference, bubble scheduling, buffer optimization, and batched metrics reduce latency by 36.04 ms, 10 ms, and 21.14 ms respectively, yielding 148.06 ms. For sparse inference, utilizing sparse kernels, memory coalescing, and kernel fusion cuts latency by 36.04 ms, 13.49 ms, and 1.76 ms, resulting in 108.24 ms.

\textbf{Trade-off between Performance and Latency. }
As shown in Fig.\ref{fig:efficient}, we analyze the trade-off between success rate and latency. While the success rate remains robust below $40\%$ pruning, it deteriorates rapidly beyond this threshold. We observe an optimal trade-off between performance and latency at a $40\%$ pruning ratio.


\textbf{Overhead of Pruning Stream. }The parallel pruning stream introduces overhead $\delta$ (e.g., SM scheduling), detailed in Tab.\ref{tab:latency_analysis}. Due to the lightweight design, $\delta$ is limited to 4.5 ms.

\textbf{Hyperparameters Studies. }
Fig.\ref{fig:para} analyzes the impact of $\alpha$ and $p$ on LIBERO-Long. High $\alpha$ over-relies on historical features, causing sparsity lag, while low $\alpha$ leads to instability. Performance peaks at $\alpha=0.7$, validating our design. Similarly, low $p$ overlooks subtle cues, whereas high $p$ causes hypersensitivity to noise. The peak success rate at $p=80\%$ confirms EcoVLA effectively filters noise while maintaining the sensitivity required for robust operation. Additional analysis is provided in the App.\ref{app:hyper}
\section{Conclusion}
We present EcoVLA, a training-free, plug-and-play adaptive pruning framework for VLA models that combines Environment-aware Adaptive Pruning (EAP) with Interleaved Inference Orchestration (I\textsuperscript{2}O) to update sparsity online with negligible overhead. We validate EcoVLA on both real robots and high-fidelity simulators.  






\nocite{langley00}

\bibliography{main}

@article{black2410pi0,
  title={$\pi$0: A vision-language-action flow model for general robot control. CoRR, abs/2410.24164, 2024. doi: 10.48550},
  author={Black, Kevin and Brown, Noah and Driess, Danny and Esmail, Adnan and Equi, Michael and Finn, Chelsea and Fusai, Niccolo and Groom, Lachy and Hausman, Karol and Ichter, Brian and others},
  journal={arXiv preprint ARXIV.2410.24164}
}

@article{brohan2022rt,
  title={Rt-1: Robotics transformer for real-world control at scale},
  author={Brohan, Anthony and Brown, Noah and Carbajal, Justice and Chebotar, Yevgen and Dabis, Joseph and Finn, Chelsea and Gopalakrishnan, Keerthana and Hausman, Karol and Herzog, Alex and Hsu, Jasmine and others},
  journal={arXiv preprint arXiv:2212.06817},
  year={2022}
}

@article{kim2024openvla,
  title={Openvla: An open-source vision-language-action model},
  author={Kim, Moo Jin and Pertsch, Karl and Karamcheti, Siddharth and Xiao, Ted and Balakrishna, Ashwin and Nair, Suraj and Rafailov, Rafael and Foster, Ethan and Lam, Grace and Sanketi, Pannag and others},
  journal={arXiv preprint arXiv:2406.09246},
  year={2024}
}

@article{kim2025fine,
  title={Fine-tuning vision-language-action models: Optimizing speed and success},
  author={Kim, Moo Jin and Finn, Chelsea and Liang, Percy},
  journal={arXiv preprint arXiv:2502.19645},
  year={2025}
}

@article{li2024cogact,
  title={Cogact: A foundational vision-language-action model for synergizing cognition and action in robotic manipulation},
  author={Li, Qixiu and Liang, Yaobo and Wang, Zeyu and Luo, Lin and Chen, Xi and Liao, Mozheng and Wei, Fangyun and Deng, Yu and Xu, Sicheng and Zhang, Yizhong and others},
  journal={arXiv preprint arXiv:2411.19650},
  year={2024}
}

@article{intelligence2504pi0,
  title={$\pi$0. 5: a vision-language-action model with open-world generalization, 2025},
  author={Intelligence, Physical and Black, Kevin and Brown, Noah and Darpinian, James and Dhabalia, Karan and Driess, Danny and Esmail, Adnan and Equi, Michael and Finn, Chelsea and Fusai, Niccolo and others},
  journal={URL https://arxiv. org/abs/2504.16054},
  volume={1},
  number={2},
  pages={3}
}

@article{xu2025vla,
  title={Vla-cache: Towards efficient vision-language-action model via adaptive token caching in robotic manipulation},
  author={Xu, Siyu and Wang, Yunke and Xia, Chenghao and Zhu, Dihao and Huang, Tao and Xu, Chang},
  journal={arXiv preprint arXiv:2502.02175},
  year={2025}
}

@article{yang2025efficientvla,
  title={EfficientVLA: Training-Free Acceleration and Compression for Vision-Language-Action Models},
  author={Yang, Yantai and Wang, Yuhao and Wen, Zichen and Zhongwei, Luo and Zou, Chang and Zhang, Zhipeng and Wen, Chuan and Zhang, Linfeng},
  journal={arXiv preprint arXiv:2506.10100},
  year={2025}
}

@article{liu2025vla,
  title={VLA-Pruner: Temporal-Aware Dual-Level Visual Token Pruning for Efficient Vision-Language-Action Inference},
  author={Liu, Ziyan and Chen, Yeqiu and Cai, Hongyi and Lin, Tao and Yang, Shuo and Liu, Zheng and Zhao, Bo},
  journal={arXiv preprint arXiv:2511.16449},
  year={2025}
}

@article{jabbour2025don,
  title={Don't Run with Scissors: Pruning Breaks VLA Models but They Can Be Recovered},
  author={Jabbour, Jason and Kim, Dong-Ki and Smith, Max and Patrikar, Jay and Ghosal, Radhika and Wang, Youhui and Agha, Ali and Reddi, Vijay Janapa and Omidshafiei, Shayegan},
  journal={arXiv preprint arXiv:2510.08464},
  year={2025}
}

@article{yue2024deer,
  title={Deer-vla: Dynamic inference of multimodal large language models for efficient robot execution},
  author={Yue, Yang and Wang, Yulin and Kang, Bingyi and Han, Yizeng and Wang, Shenzhi and Song, Shiji and Feng, Jiashi and Huang, Gao},
  journal={Advances in Neural Information Processing Systems},
  volume={37},
  pages={56619--56643},
  year={2024}
}

@article{zhang2025mole,
  title={Mole-vla: Dynamic layer-skipping vision language action model via mixture-of-layers for efficient robot manipulation},
  author={Zhang, Rongyu and Dong, Menghang and Zhang, Yuan and Heng, Liang and Chi, Xiaowei and Dai, Gaole and Du, Li and Du, Yuan and Zhang, Shanghang},
  journal={arXiv preprint arXiv:2503.20384},
  year={2025}
}

@article{wang2025specprune,
  title={Specprune-vla: Accelerating vision-language-action models via action-aware self-speculative pruning},
  author={Wang, Hanzhen and Xu, Jiaming and Pan, Jiayi and Zhou, Yongkang and Dai, Guohao},
  journal={arXiv preprint arXiv:2509.05614},
  year={2025}
}

@article{chen2025rlrc,
  title={RLRC: Reinforcement Learning-based Recovery for Compressed Vision-Language-Action Models},
  author={Chen, Yuxuan and Li, Xiao},
  journal={arXiv preprint arXiv:2506.17639},
  year={2025}
}

@inproceedings{liu2023deja,
  title={Deja vu: Contextual sparsity for efficient llms at inference time},
  author={Liu, Zichang and Wang, Jue and Dao, Tri and Zhou, Tianyi and Yuan, Binhang and Song, Zhao and Shrivastava, Anshumali and Zhang, Ce and Tian, Yuandong and Re, Christopher and others},
  booktitle={International Conference on Machine Learning},
  pages={22137--22176},
  year={2023},
  organization={PMLR}
}

@article{le2025probe,
  title={Probe pruning: Accelerating llms through dynamic pruning via model-probing},
  author={Le, Qi and Diao, Enmao and Wang, Ziyan and Wang, Xinran and Ding, Jie and Yang, Li and Anwar, Ali},
  journal={arXiv preprint arXiv:2502.15618},
  year={2025}
}

@article{black2025real,
  title={Real-Time Execution of Action Chunking Flow Policies},
  author={Black, Kevin and Galliker, Manuel Y and Levine, Sergey},
  journal={arXiv preprint arXiv:2506.07339},
  year={2025}
}

@article{ma2025running,
  title={Running vlas at real-time speed},
  author={Ma, Yunchao and Zhou, Yizhuang and Yang, Yunhuan and Wang, Tiancai and Fan, Haoqiang},
  journal={arXiv preprint arXiv:2510.26742},
  year={2025}
}

@article{wen2025tinyvla,
  title={Tinyvla: Towards fast, data-efficient vision-language-action models for robotic manipulation},
  author={Wen, Junjie and Zhu, Yichen and Li, Jinming and Zhu, Minjie and Tang, Zhibin and Wu, Kun and Xu, Zhiyuan and Liu, Ning and Cheng, Ran and Shen, Chaomin and others},
  journal={IEEE Robotics and Automation Letters},
  year={2025},
  publisher={IEEE}
}

@inproceedings{an2024fluctuation,
  title={Fluctuation-based adaptive structured pruning for large language models},
  author={An, Yongqi and Zhao, Xu and Yu, Tao and Tang, Ming and Wang, Jinqiao},
  booktitle={Proceedings of the AAAI Conference on Artificial Intelligence},
  volume={38},
  number={10},
  pages={10865--10873},
  year={2024}
}

@article{sun2023simple,
  title={A simple and effective pruning approach for large language models},
  author={Sun, Mingjie and Liu, Zhuang and Bair, Anna and Kolter, J Zico},
  journal={arXiv preprint arXiv:2306.11695},
  year={2023}
}

@inproceedings{chen2024image,
  title={An image is worth 1/2 tokens after layer 2: Plug-and-play inference acceleration for large vision-language models},
  author={Chen, Liang and Zhao, Haozhe and Liu, Tianyu and Bai, Shuai and Lin, Junyang and Zhou, Chang and Chang, Baobao},
  booktitle={European Conference on Computer Vision},
  pages={19--35},
  year={2024},
  organization={Springer}
}

@article{dao2022flashattention,
  title={Flashattention: Fast and memory-efficient exact attention with io-awareness},
  author={Dao, Tri and Fu, Dan and Ermon, Stefano and Rudra, Atri and R{\'e}, Christopher},
  journal={Advances in neural information processing systems},
  volume={35},
  pages={16344--16359},
  year={2022}
}

@article{liu2023libero,
  title={Libero: Benchmarking knowledge transfer for lifelong robot learning},
  author={Liu, Bo and Zhu, Yifeng and Gao, Chongkai and Feng, Yihao and Liu, Qiang and Zhu, Yuke and Stone, Peter},
  journal={Advances in Neural Information Processing Systems},
  volume={36},
  pages={44776--44791},
  year={2023}
}

@article{li2024evaluating,
  title={Evaluating real-world robot manipulation policies in simulation},
  author={Li, Xuanlin and Hsu, Kyle and Gu, Jiayuan and Pertsch, Karl and Mees, Oier and Walke, Homer Rich and Fu, Chuyuan and Lunawat, Ishikaa and Sieh, Isabel and Kirmani, Sean and others},
  journal={arXiv preprint arXiv:2405.05941},
  year={2024}
}

@article{jin2022pruning,
  title={Pruning’s effect on generalization through the lens of training and regularization},
  author={Jin, Tian and Carbin, Michael and Roy, Dan and Frankle, Jonathan and Dziugaite, Gintare Karolina},
  journal={Advances in Neural Information Processing Systems},
  volume={35},
  pages={37947--37961},
  year={2022}
}

@article{shinde2025survey,
  title={A Survey on Efficient Vision-Language Models},
  author={Shinde, Gaurav and Ravi, Anuradha and Dey, Emon and Sakib, Shadman and Rampure, Milind and Roy, Nirmalya},
  journal={Wiley Interdisciplinary Reviews: Data Mining and Knowledge Discovery},
  volume={15},
  number={3},
  pages={e70036},
  year={2025},
  publisher={Wiley Online Library}
}

\bibliographystyle{icml2026}

\newpage
\appendix
\onecolumn

\section{Experimental Settings}
\label{appendix:Baseline}
\subsection{VLA Model Details}
\textbf{OpenVLA-OFT. }Derived from the OpenVLA architecture, OpenVLA-OFT incorporates an Optimized Fine-Tuning strategy that simultaneously enhances manipulation precision and computational efficiency. By adopting a continuous action space optimized via $L1$ regression alongside parallel decoding and action chunking mechanisms, it achieves superior throughput and success rates on the LIBERO benchmark. A key characteristic of this variant is the deployment of bidirectional attention during inference. 

\textbf{$\pi_{0.5}$. }$\pi_{0.5}$ is a generalist Vision-Language-Action (VLA) model designed to achieve broad open-world generalization in mobile manipulation by leveraging a heterogeneous co-training recipe. Building upon the $\pi_{0}$ architecture and the PaliGemma VLM backbone, $\pi_{0.5}$ integrates diverse data sources—including cross-embodiment robot data, multimodal web data, and high-level semantic predictions—to bridge the generalization gap. The model employs a hierarchical inference mechanism where it first predicts a high-level semantic subtask (e.g., "pick up the pillow") based on the visual observation and global instruction, which then conditions a specialized "action expert" to generate continuous low-level control actions via flow matching. This capability is developed through a two-stage training process that transitions from scalable discrete token pre-training to precise continuous flow-matching post-training, enabling the execution of complex, long-horizon tasks in environments completely unseen during training.

\textbf{CogACT. }This architecture synthesizes perception and reasoning by employing DINOv2 and SigLIP for visual encoding alongside a Llama2-7B language backbone. To bridge the gap between high-level cognition and low-level action, CogACT utilizes a specialized Diffusion Transformer (DiT). By conditioning this diffusion-based action module on the features extracted by the VLM, the model effectively addresses the challenges of generating precise, continuous, and temporally correlated robotic trajectories.

\subsection{Acceleration Method Details}
\textbf{FastV. }FastV addresses inference latency in Large Vision-Language Models (LVLMs) by mitigating visual token redundancy. The method is grounded in the observation that deep layers often exhibit an attention sink phenomenon, where visual tokens consume substantial computational resources despite receiving minimal attention weights. To counter this, FastV introduces a plug-and-play mechanism that monitors attention scores to dynamically discard low-utility visual tokens after a certain depth, thereby reducing FLOPs while preserving model accuracy.

\textbf{VLA-Cache. }VLA-Cache is a training-free inference accelerator tailored for robotic VLA models. It exploits the observation that visual scenes in robotic tasks remain largely stable between consecutive frames, particularly in background areas. By distinguishing between static and dynamic elements, the method recycles KV-cache states for unchanged tokens while enforcing full computation for critical, task-relevant features to ensure precision. Furthermore, it incorporates an adaptive strategy that modulates the caching ratio according to layer-specific attention patterns.

\subsection{Implementation Details}
\label{appendix:Implementation}
For OpenVLA-OFT, we set the sensitivity parameter $p=5$. Regarding temporal inertia parameter, we set $\alpha=0.7$ for LIBERO-Spatial and LIBERO-Long to reduce historical reliance in dynamic spatial layouts. Conversely, for LIBERO-Object and LIBERO-Goal, we use a higher $\alpha=0.9$ as these manipulation tasks feature relatively stable environments. 
For $\pi_{0.5}$, given its enhanced execution stability, we set the sensitivity parameter to 80 to mitigate environmental noise. The temporal inertia parameter is kept fixed at 0.7. For CogACT, we set the sensitivity parameter $p=5$, $\alpha=0.7$. 

\begin{figure}[h]

    \centering
    \includegraphics[width=0.4\textwidth]{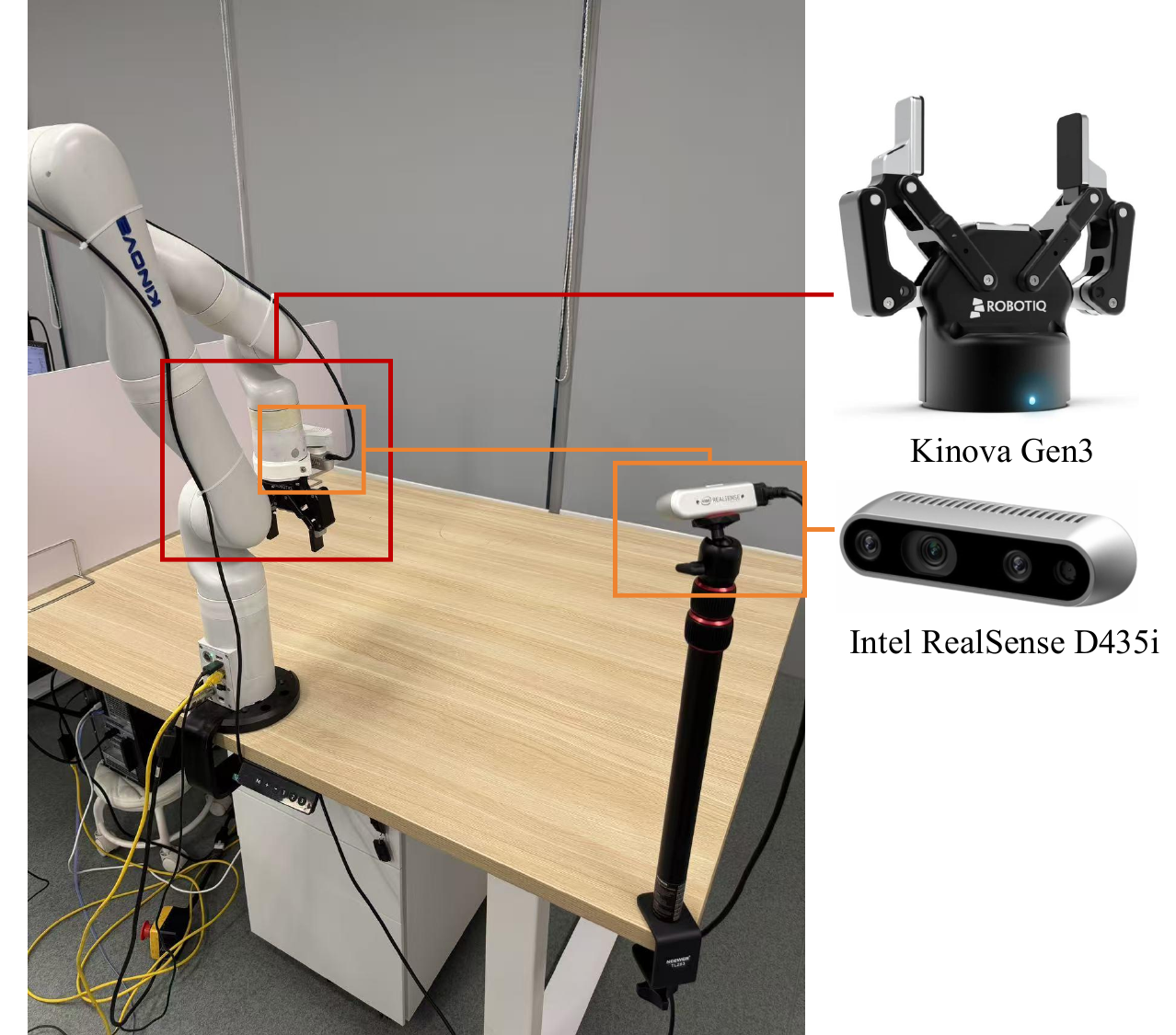}
    \caption{Kinova Gen3 Robot Setup.}
    \label{fig:realrobotpic}
    \vskip -1.5em
\end{figure}

\subsection{Benchmarks Details}
\label{app:bench}
\textbf{Real-Robot Setup. } We conducted experiments on a 7-DoF Kinova Gen3 robotic arm, with the hardware setup shown in Fig.\ref{fig:realrobotpic}. Our system uses two Intel RealSense D435i cameras: one providing a third-person view and the other mounted on the wrist.

\textbf{Training Details. } For each task, we collected $50$ demonstrations. When fine-tuning the $\pi_{0.5}$ model, we froze the VLM backbone and performed LoRA-based adaptation only, while fullly fine-tuning the action expert. We used a global batch size of $32$ and trained for $10k$ steps on a single GPU.

\section{Real-robot Analysis}
\label{app:realrobot}
We conduct experiments on a real-robot platform. Specifically, we fine-tune $\pi_{0.5}$ on our collected dataset, training each task for 10,000 steps with a batch size of 32. During evaluation, to more thoroughly assess the model’s capability, we test on 20 randomly sampled object placements within the training range, re-randomizing the object position in every trial to probe the model’s spatial reasoning and generalization across object locations. On real-robot tasks, while achieving a $1.26\times$ speedup, EcoVLA exhibits a minor performance drop compared to $\pi_{0.5}$. We further observe that most failures occur when the object is placed near the boundary of the workspace. We conjecture that such edge-case placements require finer-grained spatial cues and stricter geometric constraints, making the pruned model more susceptible to representation loss and thus less robust in spatial generalization.

\section{Hyperparameter Analysis}
\label{app:hyper}

We evaluate the influence of sensitivity parameter $p$ and temporal inertia parameter, $\alpha$ on success rates using $\pi_{0.5}$.

\textbf{Impact of Sensitivity Parameter. }
As shown in Fig.\ref{fig:para} (b), we evaluate the impact of the sensitivity parameter $p$, as it directly dictates EcoVLA's responsiveness to environmental dynamics. At low $p$ values, EcoVLA adopts a conservative strategy, updating sparsity patterns only during drastic environmental changes, thereby overlooking subtle cues. Conversely, an excessively high $p$ induces hypersensitivity, where sensor noise or lighting fluctuations are misidentified as dynamics. This triggers redundant updates that disrupt temporal continuity. Performance peaks at $p=80\%$, a setting that effectively filters noise while retaining sufficient sensitivity to capture physical transitions, ensuring robust operation in complex environments.

\textbf{Impact of Temporal Inertia Parameter. }
We investigate the properties of $\alpha$ on LIBERO-Long (Fig.\ref{fig:para} (a)), as this parameter is critical for long-horizon tasks with varying spatial layouts. High $\alpha$ excessively weights history, causing the model to miss new critical features, whereas extremely low $\alpha$ relies on single-frame inputs, disrupting inference stability. Performance peaks at $87.0\%$ with an intermediate $\alpha$, validating our Temporal Consistency Pruning strategy. This confirms that EcoVLA effectively filters high-frequency noise while maintaining sensitivity to environmental dynamics.

\end{document}